\documentclass{edm_template}  
\usepackage{amsmath,epsfig} 
\usepackage{amsfonts}
\usepackage{amssymb}
\usepackage{mathrsfs}
\usepackage{subfigure}
\usepackage{paralist}
\usepackage{booktabs}
\usepackage{color}
\DeclareGraphicsExtensions{.pdf,.jpeg,.png,.epbibdesks}
\usepackage[T1]{fontenc}
\usepackage{epstopdf}
\usepackage{graphicx}

\usepackage{floatrow}
\newfloatcommand{capbtabbox}{table}[][\FBwidth]
\usepackage[algo2e]{algorithm2e}
\usepackage{yhmath}
\usepackage{amssymb}
\usepackage{amsfonts}
\usepackage{mathrsfs}
\usepackage{xspace}
\usepackage{bm}
\usepackage{upgreek}

\newcommand{\safemath}[2]{\newcommand{#1}{\ensuremath{#2}\xspace}}

\safemath{\bma}{\mathbf{a}}
\safemath{\bmb}{\mathbf{b}}
\safemath{\bmc}{\mathbf{c}}
\safemath{\bmd}{\mathbf{d}}
\safemath{\bme}{\mathbf{e}}
\safemath{\bmf}{\mathbf{f}}
\safemath{\bmg}{\mathbf{g}}
\safemath{\bmh}{\mathbf{h}}
\safemath{\bmi}{\mathbf{i}}
\safemath{\bmj}{\mathbf{j}}
\safemath{\bmk}{\mathbf{k}}
\safemath{\bml}{\mathbf{l}}
\safemath{\bmm}{\mathbf{m}}
\safemath{\bmn}{\mathbf{n}}
\safemath{\bmo}{\mathbf{o}}
\safemath{\bmp}{\mathbf{p}}
\safemath{\bmq}{\mathbf{q}}
\safemath{\bmr}{\mathbf{r}}
\safemath{\bms}{\mathbf{s}}
\safemath{\bmt}{\mathbf{t}}
\safemath{\bmu}{\mathbf{u}}
\safemath{\bmv}{\mathbf{v}}
\safemath{\bmw}{\mathbf{w}}
\safemath{\bmx}{\mathbf{x}}
\safemath{\bmy}{\mathbf{y}}
\safemath{\bmz}{\mathbf{z}}
\safemath{\bmzero}{\mathbf{0}}
\safemath{\bmone}{\mathbf{1}}

\bmdefine{\biad}{a}
\bmdefine{\bibd}{b}
\bmdefine{\bicd}{c}
\bmdefine{\bidd}{d}
\bmdefine{\bied}{e}
\bmdefine{\bifd}{f}
\bmdefine{\bigd}{g}
\bmdefine{\bihd}{h}
\bmdefine{\biid}{i}
\bmdefine{\bijd}{j}
\bmdefine{\bikd}{k}
\bmdefine{\bild}{l}
\bmdefine{\bimd}{m}
\bmdefine{\bind}{n}
\bmdefine{\biod}{o}
\bmdefine{\bipd}{p}
\bmdefine{\biqd}{q}
\bmdefine{\bird}{r}
\bmdefine{\bisd}{s}
\bmdefine{\bitd}{t}
\bmdefine{\biud}{u}
\bmdefine{\bivd}{v}
\bmdefine{\biwd}{w}
\bmdefine{\bixd}{x}
\bmdefine{\biyd}{y}
\bmdefine{\bizd}{z}

\bmdefine{\bixid}{\xi}
\bmdefine{\bilambdad}{\lambda}
\bmdefine{\bimud}{\mu}
\bmdefine{\bithetad}{\theta}
\bmdefine{\biphid}{\phi}
\bmdefine{\bideltad}{\delta}

\safemath{\bmia}{\biad}
\safemath{\bmib}{\bibd}
\safemath{\bmic}{\bicd}
\safemath{\bmid}{\bidd}
\safemath{\bmie}{\bied}
\safemath{\bmif}{\bifd}
\safemath{\bmig}{\bigd}
\safemath{\bmih}{\bihd}
\safemath{\bmii}{\biid}
\safemath{\bmij}{\bijd}
\safemath{\bmik}{\bikd}
\safemath{\bmil}{\bild}
\safemath{\bmim}{\bimd}
\safemath{\bmin}{\bind}
\safemath{\bmio}{\biod}
\safemath{\bmip}{\bipd}
\safemath{\bmiq}{\biqd}
\safemath{\bmir}{\bird}
\safemath{\bmis}{\bisd}
\safemath{\bmit}{\bitd}
\safemath{\bmiu}{\biud}
\safemath{\bmiv}{\bivd}
\safemath{\bmiw}{\biwd}
\safemath{\bmix}{\bixd}
\safemath{\bmiy}{\biyd}
\safemath{\bmiz}{\bizd}

\safemath{\bmxi}{\bixid}
\safemath{\bmlambda}{\bilambdad}
\safemath{\bmmu}{\bimud}
\safemath{\bmtheta}{\bithetad}
\safemath{\bmphi}{\biphid}
\safemath{\bmdelta}{\bideltad}

\safemath{\bA}{\mathbf{A}}
\safemath{\bB}{\mathbf{B}}
\safemath{\bC}{\mathbf{C}}
\safemath{\bD}{\mathbf{D}}
\safemath{\bE}{\mathbf{E}}
\safemath{\bF}{\mathbf{F}}
\safemath{\bG}{\mathbf{G}}
\safemath{\bH}{\mathbf{H}}
\safemath{\bI}{\mathbf{I}}
\safemath{\bJ}{\mathbf{J}}
\safemath{\bK}{\mathbf{K}}
\safemath{\bL}{\mathbf{L}}
\safemath{\bM}{\mathbf{M}}
\safemath{\bN}{\mathbf{N}}
\safemath{\bO}{\mathbf{O}}
\safemath{\bP}{\mathbf{P}}
\safemath{\bQ}{\mathbf{Q}}
\safemath{\bR}{\mathbf{R}}
\safemath{\bS}{\mathbf{S}}
\safemath{\bT}{\mathbf{T}}
\safemath{\bU}{\mathbf{U}}
\safemath{\bV}{\mathbf{V}}
\safemath{\bW}{\mathbf{W}}
\safemath{\bX}{\mathbf{X}}
\safemath{\bY}{\mathbf{Y}}
\safemath{\bZ}{\mathbf{Z}}

\safemath{\bZero}{\mathbf{0}}
\safemath{\bOne}{\mathbf{1}}
\safemath{\bDelta}{\mathbf{\Delta}}
\safemath{\bLambda}{\mathbf{\UpLambda}}
\safemath{\bPhi}{\mathbf{\Upphi}}
\safemath{\bSigma}{\mathbf{\Upsigma}}
\safemath{\bOmega}{\mathbf{\Upomega}}
\safemath{\bTheta}{\mathbf{\Uptheta}}

\bmdefine{\biAd}{A}
\bmdefine{\biBd}{B}
\bmdefine{\biCd}{C}
\bmdefine{\biDd}{D}
\bmdefine{\biEd}{E}
\bmdefine{\biFd}{F}
\bmdefine{\biGd}{G}
\bmdefine{\biHd}{H}
\bmdefine{\biId}{I}
\bmdefine{\biJd}{J}
\bmdefine{\biKd}{K}
\bmdefine{\biLd}{L}
\bmdefine{\biMd}{M}
\bmdefine{\biNd}{N}
\bmdefine{\biOd}{O}
\bmdefine{\biPd}{P}
\bmdefine{\biQd}{Q}
\bmdefine{\biRd}{R}
\bmdefine{\biSd}{S}
\bmdefine{\biTd}{T}
\bmdefine{\biUd}{U}
\bmdefine{\biVd}{V}
\bmdefine{\biWd}{W}
\bmdefine{\biXd}{X}
\bmdefine{\biYd}{Y}
\bmdefine{\biZd}{Z}

\bmdefine{\biDelta}{\Delta}
\bmdefine{\biLambda}{\Lambda}
\bmdefine{\biPhi}{\Phi}
\bmdefine{\biSigma}{\Sigma}
\bmdefine{\biOmega}{\Omega}
\bmdefine{\biTheta}{\Theta}

\safemath{\bimA}{\biAd}
\safemath{\bimB}{\biBd}
\safemath{\bimC}{\biCd}
\safemath{\bimD}{\biDd}
\safemath{\bimE}{\biEd}
\safemath{\bimF}{\biFd}
\safemath{\bimG}{\biGd}
\safemath{\bimH}{\biHd}
\safemath{\bimI}{\biId}
\safemath{\bimJ}{\biJd}
\safemath{\bimK}{\biKd}
\safemath{\bimL}{\biLd}
\safemath{\bimM}{\biMd}
\safemath{\bimN}{\biNd}
\safemath{\bimO}{\biOd}
\safemath{\bimP}{\biPd}
\safemath{\bimQ}{\biQd}
\safemath{\bimR}{\biRd}
\safemath{\bimS}{\biSd}
\safemath{\bimT}{\biTd}
\safemath{\bimU}{\biUd}
\safemath{\bimV}{\biVd}
\safemath{\bimW}{\biWd}
\safemath{\bimX}{\biXd}
\safemath{\bimY}{\biYd}
\safemath{\bimZ}{\biZd}

\safemath{\bimDelta}{\biDelta}
\safemath{\bimLambda}{\biLambda}
\safemath{\bimPhi}{\biPhi}
\safemath{\bimSigma}{\biSigma}
\safemath{\bimOmega}{\biOmega}
\safemath{\bimTheta}{\biTheta}

\safemath{\setA}{\mathcal{A}}
\safemath{\setB}{\mathcal{B}}
\safemath{\setC}{\mathcal{C}}
\safemath{\setD}{\mathcal{D}}
\safemath{\setE}{\mathcal{E}}
\safemath{\setF}{\mathcal{F}}
\safemath{\setG}{\mathcal{G}}
\safemath{\setH}{\mathcal{H}}
\safemath{\setI}{\mathcal{I}}
\safemath{\setJ}{\mathcal{J}}
\safemath{\setK}{\mathcal{K}}
\safemath{\setL}{\mathcal{L}}
\safemath{\setM}{\mathcal{M}}
\safemath{\setN}{\mathcal{N}}
\safemath{\setO}{\mathcal{O}}
\safemath{\setP}{\mathcal{P}}
\safemath{\setQ}{\mathcal{Q}}
\safemath{\setR}{\mathcal{R}}
\safemath{\setS}{\mathcal{S}}
\safemath{\setT}{\mathcal{T}}
\safemath{\setU}{\mathcal{U}}
\safemath{\setV}{\mathcal{V}}
\safemath{\setW}{\mathcal{W}}
\safemath{\setX}{\mathcal{X}}
\safemath{\setY}{\mathcal{Y}}
\safemath{\setZ}{\mathcal{Z}}
\safemath{\emptySet}{\varnothing}

\safemath{\colA}{\mathscr{A}}
\safemath{\colB}{\mathscr{B}}
\safemath{\colC}{\mathscr{C}}
\safemath{\colD}{\mathscr{D}}
\safemath{\colE}{\mathscr{E}}
\safemath{\colF}{\mathscr{F}}
\safemath{\colG}{\mathscr{G}}
\safemath{\colH}{\mathscr{H}}
\safemath{\colI}{\mathscr{I}}
\safemath{\colJ}{\mathscr{J}}
\safemath{\colK}{\mathscr{K}}
\safemath{\colL}{\mathscr{L}}
\safemath{\colM}{\mathscr{M}}
\safemath{\colN}{\mathscr{N}}
\safemath{\colO}{\mathscr{O}}
\safemath{\colP}{\mathscr{P}}
\safemath{\colQ}{\mathscr{Q}}
\safemath{\colR}{\mathscr{R}}
\safemath{\colS}{\mathscr{S}}
\safemath{\colT}{\mathscr{T}}
\safemath{\colU}{\mathscr{U}}
\safemath{\colV}{\mathscr{V}}
\safemath{\colW}{\mathscr{W}}
\safemath{\colX}{\mathscr{X}}
\safemath{\colY}{\mathscr{Y}}
\safemath{\colZ}{\mathscr{Z}}

\safemath{\opA}{\mathbb{A}}
\safemath{\opB}{\mathbb{B}}
\safemath{\opC}{\mathbb{C}}
\safemath{\opD}{\mathbb{D}}
\safemath{\opE}{\mathbb{E}}
\safemath{\opF}{\mathbb{F}}
\safemath{\opG}{\mathbb{G}}
\safemath{\opH}{\mathbb{H}}
\safemath{\opI}{\mathbb{I}}
\safemath{\opJ}{\mathbb{J}}
\safemath{\opK}{\mathbb{K}}
\safemath{\opL}{\mathbb{L}}
\safemath{\opM}{\mathbb{M}}
\safemath{\opN}{\mathbb{N}}
\safemath{\opO}{\mathbb{O}}
\safemath{\opP}{\mathbb{P}}
\safemath{\opQ}{\mathbb{Q}}
\safemath{\opR}{\mathbb{R}}
\safemath{\opS}{\mathbb{S}}
\safemath{\opT}{\mathbb{T}}
\safemath{\opU}{\mathbb{U}}
\safemath{\opV}{\mathbb{V}}
\safemath{\opW}{\mathbb{W}}
\safemath{\opX}{\mathbb{X}}
\safemath{\opY}{\mathbb{Y}}
\safemath{\opZ}{\mathbb{Z}}
\safemath{\opZero}{\mathbb{O}}
\safemath{\identityop}{\opI}

\safemath{\veca}{\bma}
\safemath{\vecb}{\bmb}
\safemath{\vecc}{\bmc}
\safemath{\vecd}{\bmd}
\safemath{\vece}{\bme}
\safemath{\vecf}{\bmf}
\safemath{\vecg}{\bmg}
\safemath{\vech}{\bmh}
\safemath{\veci}{\bmi}
\safemath{\vecj}{\bmj}
\safemath{\veck}{\bmk}
\safemath{\vecl}{\bml}
\safemath{\vecm}{\bmm}
\safemath{\vecn}{\bmn}
\safemath{\veco}{\bmo}
\safemath{\vecp}{\bmp}
\safemath{\vecq}{\bmq}
\safemath{\vecr}{\bmr}
\safemath{\vecs}{\bms}
\safemath{\vect}{\bmt}
\safemath{\vecu}{\bmu}
\safemath{\vecv}{\bmv}
\safemath{\vecw}{\bmw}
\safemath{\vecx}{\bmx}
\safemath{\vecy}{\bmy}
\safemath{\vecz}{\bmz}

\safemath{\veczero}{\bmzero}
\safemath{\vecone}{\bmone}
\safemath{\vecxi}{\bmxi}
\safemath{\veclambda}{\bmlambda}
\safemath{\vecmu}{\bmmu}
\safemath{\vectheta}{\bmtheta}
\safemath{\vecphi}{\bmphi}
\safemath{\vecdelta}{\bmdelta}

\safemath{\matA}{\bA}
\safemath{\matB}{\bB}
\safemath{\matC}{\bC}
\safemath{\matD}{\bD}
\safemath{\matE}{\bE}
\safemath{\matF}{\bF}
\safemath{\matG}{\bG}
\safemath{\matH}{\bH}
\safemath{\matI}{\bI}
\safemath{\matJ}{\bJ}
\safemath{\matK}{\bK}
\safemath{\matL}{\bL}
\safemath{\matM}{\bM}
\safemath{\matN}{\bN}
\safemath{\matO}{\bO}
\safemath{\matP}{\bP}
\safemath{\matQ}{\bQ}
\safemath{\matR}{\bR}
\safemath{\matS}{\bS}
\safemath{\matT}{\bT}
\safemath{\matU}{\bU}
\safemath{\matV}{\bV}
\safemath{\matW}{\bW}
\safemath{\matX}{\bX}
\safemath{\matY}{\bY}
\safemath{\matZ}{\bZ}
\safemath{\matzero}{\bmzero}

\safemath{\matDelta}{\bDelta}
\safemath{\matLambda}{\bLambda}
\safemath{\matPhi}{\bPhi}
\safemath{\matSigma}{\bSigma}
\safemath{\matOmega}{\bOmega}
\safemath{\matTheta}{\bTheta}

\safemath{\matidentity}{\matI}
\safemath{\matone}{\matO}

\safemath{\rnda}{A}
\safemath{\rndb}{B}
\safemath{\rndc}{C}
\safemath{\rndd}{D}
\safemath{\rnde}{E}
\safemath{\rndf}{F}
\safemath{\rndg}{G}
\safemath{\rndh}{H}
\safemath{\rndi}{I}
\safemath{\rndj}{J}
\safemath{\rndk}{K}
\safemath{\rndl}{L}
\safemath{\rndm}{M}
\safemath{\rndn}{N}
\safemath{\rndo}{O}
\safemath{\rndp}{P}
\safemath{\rndq}{Q}
\safemath{\rndr}{R}
\safemath{\rnds}{S}
\safemath{\rndt}{T}
\safemath{\rndu}{U}
\safemath{\rndv}{V}
\safemath{\rndw}{W}
\safemath{\rndx}{X}
\safemath{\rndy}{Y}
\safemath{\rndz}{Z}

\safemath{\rveca}{\bimA}
\safemath{\rvecb}{\bimB}
\safemath{\rvecc}{\bimC}
\safemath{\rvecd}{\bimD}
\safemath{\rvece}{\bimE}
\safemath{\rvecf}{\bimF}
\safemath{\rvecg}{\bimG}
\safemath{\rvech}{\bimH}
\safemath{\rveci}{\bimI}
\safemath{\rvecj}{\bimJ}
\safemath{\rveck}{\bimK}
\safemath{\rvecl}{\bimL}
\safemath{\rvecm}{\bimM}
\safemath{\rvecn}{\bimN}
\safemath{\rveco}{\bomO}
\safemath{\rvecp}{\bimP}
\safemath{\rvecq}{\bimQ}
\safemath{\rvecr}{\bimR}
\safemath{\rvecs}{\bimS}
\safemath{\rvect}{\bimT}
\safemath{\rvecu}{\bimU}
\safemath{\rvecv}{\bimV}
\safemath{\rvecw}{\bimW}
\safemath{\rvecx}{\bimX}
\safemath{\rvecy}{\bimY}
\safemath{\rvecz}{\bimZ}

\safemath{\rvecxi}{\bmxi}
\safemath{\rveclambda}{\bmlambda}
\safemath{\rvecmu}{\bmmu}
\safemath{\rvectheta}{\bmtheta}
\safemath{\rvecphi}{\bmphi}

\safemath{\rmatA}{\bimA}
\safemath{\rmatB}{\bimB}
\safemath{\rmatC}{\bimC}
\safemath{\rmatD}{\bimD}
\safemath{\rmatE}{\bimE}
\safemath{\rmatF}{\bimF}
\safemath{\rmatG}{\bimG}
\safemath{\rmatH}{\bimH}
\safemath{\rmatI}{\bimI}
\safemath{\rmatJ}{\bimJ}
\safemath{\rmatK}{\bimK}
\safemath{\rmatL}{\bimL}
\safemath{\rmatM}{\bimM}
\safemath{\rmatN}{\bimN}
\safemath{\rmatO}{\bimO}
\safemath{\rmatP}{\bimP}
\safemath{\rmatQ}{\bimQ}
\safemath{\rmatR}{\bimR}
\safemath{\rmatS}{\bimS}
\safemath{\rmatT}{\bimT}
\safemath{\rmatU}{\bimU}
\safemath{\rmatV}{\bimV}
\safemath{\rmatW}{\bimW}
\safemath{\rmatX}{\bimX}
\safemath{\rmatY}{\bimY}
\safemath{\rmatZ}{\bimZ}

\safemath{\rmatDelta}{\bimDelta}
\safemath{\rmatLambda}{\bimLambda}
\safemath{\rmatPhi}{\bimPhi}
\safemath{\rmatSigma}{\bimSigma}
\safemath{\rmatOmega}{\bimOmega}
\safemath{\rmatTheta}{\bimTheta}

\usepackage{amssymb}
\usepackage{amsfonts}
\usepackage{mathrsfs}
\usepackage{xspace}
\usepackage{bm}
\usepackage{fancyref}
\usepackage{textcomp}

\usepackage{multirow}
\usepackage{stmaryrd}

\newenvironment{textbmatrix}{	\setlength{\arraycolsep}{2.5pt}%
								\big[\begin{matrix}}{\end{matrix}\big]%
								\raisebox{0.08ex}{\vphantom{M}}}

\def\be{\begin{equation}}
\def\ee{\end{equation}}
\def\een{\nonumber \end{equation}}
\def\mat{\begin{bmatrix}}
\def\emat{\end{bmatrix}}
\def\btm{\begin{textbmatrix}}
\def\etm{\end{textbmatrix}}

\def\ba#1\ea{\begin{align}#1\end{align}}
\def\bas#1\eas{\begin{align*}#1\end{align*}}
\def\bs#1\es{\begin{split}#1\end{split}} 
\def\bg#1\eg{\begin{gather}#1\end{gather}}
\def\bml#1\eml{\begin{multline}#1\end{multline}}
\def\bi#1\ei{\begin{itemize}#1\end{itemize}}

\newcommand{\lefto}{\mathopen{}\left}

\newcommand{\vecnorm}[1]{\lefto\lVert#1\right\rVert}		

\safemath{\dirac}{\delta}					
\safemath{\krond}{\dirac}					

\safemath{\upto}{\uparrow}
\safemath{\downto}{\downarrow}
\safemath{\iu}{j}							
\safemath{\ev}{\lambda}						
\safemath{\hilseqspace}{l^{2}}				
\newcommand{\banachfunspace}[1]{\setL^{#1}}	
\safemath{\hilfunspace}{\banachfunspace{2}}	

\safemath{\SNR}{\text{\sc snr}} 				
\safemath{\No}{N_0}							
\safemath{\Es}{E_s}							
\safemath{\Eb}{E_b}							
\safemath{\EbNo}{\frac{\Eb}{\No}}
\safemath{\EsNo}{\frac{\Es}{\No}}

\DeclareMathOperator{\CHop}{\ensuremath{\opH}} 
\safemath{\tvir}{\rndh_{\CHop}}				
\safemath{\tvtf}{\rndl_{\CHop}}				
\safemath{\spf}{\rnds_{\CHop}}				
\safemath{\bff}{H_{\CHop}}					

\safemath{\ircf}{r_{h}}						
\safemath{\tftvcf}{r_{s}}					
\safemath{\tfcf}{r_{l}}						
\safemath{\bfcf}{r_{H}}						

\safemath{\tcorr}{c_h}						
\safemath{\scf}{c_{s}}						
\safemath{\tfcorr}{c_{l}}					
\safemath{\fcorr}{c_{H}}						

\safemath{\mi}{I}							
\safemath{\capacity}{C}						

\safemath{\normal}{\mathcal{N}}			
\safemath{\jpg}{\mathcal{CN}}			
\safemath{\mchain}{\leftrightarrow}		

\safemath{\dB}{\,\mathrm{dB}}
\safemath{\dBm}{\,\mathrm{dBm}}
\safemath{\Hz}{\,\mathrm{Hz}}
\safemath{\kHz}{\,\mathrm{kHz}}
\safemath{\MHz}{\,\mathrm{MHz}}
\safemath{\GHz}{\,\mathrm{GHz}}
\safemath{\s}{\,\mathrm{s}}
\safemath{\ms}{\,\mathrm{ms}}
\safemath{\mus}{\,\mathrm{\text{\textmu}s}}
\safemath{\ns}{\,\mathrm{ns}}
\safemath{\ps}{\,\mathrm{ps}}
\safemath{\meter}{\,\mathrm{m}}
\safemath{\mm}{\,\mathrm{mm}}
\safemath{\cm}{\,\mathrm{cm}}
\safemath{\m}{\,\mathrm{m}}
\safemath{\W}{\,\mathrm{W}}
\safemath{\mW}{\, \mathrm{mW}}
\safemath{\J}{\,\mathrm{J}}
\safemath{\K}{\,\mathrm{K}}
\safemath{\bit}{\,\mathrm{bit}}
\safemath{\nat}{\,\mathrm{nat}}

\safemath{\define}{\triangleq}			

\safemath{\equivalent}{\sim}
\safemath{\distas}{\sim}					
\safemath{\sdiff}{\Delta}				

\safemath{\reals}{\mathbb{R}}
\safemath{\positivereals}{\reals_{+}}
\safemath{\integers}{\mathbb{Z}}
\safemath{\posint}{\integers_{+}}
\safemath{\naturals}{\mathbb{N}}
\safemath{\posnaturals}{\naturals_{+}}
\safemath{\complexset}{\mathbb{C}}
\safemath{\rationals}{\mathbb{Q}}

\newcommand*{\fancyrefapplabelprefix}{app}		
\newcommand*{\fancyrefthmlabelprefix}{thm}		
\newcommand*{\fancyreflemlabelprefix}{lem}		
\newcommand*{\fancyrefcorlabelprefix}{cor}		
\newcommand*{\fancyrefdeflabelprefix}{def}		
\newcommand*{\fancyrefalglabelprefix}{alg}		
\newcommand*{\fancyrefproplabelprefix}{prop}		
\newcommand*{\fancyrefexmpllabelprefix}{exmpl}
\newcommand*{\fancyreftbllabelprefix}{tbl}
\frefformat{vario}{\fancyrefseclabelprefix}{Section~#1}
\frefformat{vario}{\fancyrefthmlabelprefix}{Theorem~#1}
\frefformat{vario}{\fancyreflemlabelprefix}{Lemma~#1}
\frefformat{vario}{\fancyrefcorlabelprefix}{Corrolary~#1}
\frefformat{vario}{\fancyrefdeflabelprefix}{Definition~#1}
\frefformat{vario}{\fancyrefalglabelprefix}{Algorithm~#1}
\frefformat{vario}{\fancyreffiglabelprefix}{Figure~#1}
\frefformat{vario}{\fancyrefapplabelprefix}{Appendix~#1} 
\frefformat{vario}{\fancyrefeqlabelprefix}{(#1)}
\frefformat{vario}{\fancyrefproplabelprefix}{Proposition~#1}
\frefformat{vario}{\fancyrefexmpllabelprefix}{Example~#1}
\frefformat{vario}{\fancyreftbllabelprefix}{Table~#1}

\safemath{\dictab}{[\,\dicta\,\,\dictb\,]}

\safemath{\ysig}{\bmy}
\safemath{\ysighat}{\hat{\ysig}}
\safemath{\ysigdim}{M}
\safemath{\xsig}{\bmx}
\safemath{\xsigdim}{N}
\safemath{\nx}{n_x}
\safemath{\zsig}{\bmz}
\safemath{\zsigdim}{\ysigdim}
\safemath{\rsig}{\bmr}
\safemath{\Adict}{\bA}
\safemath{\Adicttilde}{\widetilde{\Adict}}
\safemath{\Adictdim}{\outputdim\times\xsigdim}
\safemath{\avec}{\bma}
\safemath{\avectilde}{\tilde{\avec}}
\safemath{\Bdict}{\bB}
\safemath{\Bdicttilde}{\widetilde{\Bdict}}
\safemath{\Cdict}{\bC}
\safemath{\cvec}{\bmc}
\safemath{\Ddict}{\bD}
\safemath{\Ddictdim}{\ysigdim\times\xsigdim}
\safemath{\dvec}{\bmd}
\safemath{\Ddicttilde}{\widetilde{\bD}}
\safemath{\Bonb}{\bB}
\safemath{\bvec}{\bmb}
\safemath{\Bonbdim}{\ysigdim\times\ysigdim}
\safemath{\noise}{\bmn}
\safemath{\noisedim}{\ysigim}
\safemath{\err}{\bme}
\safemath{\errdim}{\ysigdim}
\safemath{\errset}{\setE}
\safemath{\nerr}{n_e}
\safemath{\delop}{\bP_\errset}
\safemath{\delopc}{\bP_{{\errset}^c}}

\safemath{\cplxi}{\imath}
\safemath{\cplxj}{\jmath}

\safemath{\dict}{\matD}
\safemath{\inputdim}{N}		
\safemath{\outputdim}{M}		
\safemath{\sparsity}{S}	
\safemath{\inputdimA}{{N_a}}	
\safemath{\inputdimB}{{N_b}}	
\safemath{\elemA}{{n_a}}	
\safemath{\elemB}{{n_b}}	
\safemath{\resA}{\matR_a}	
\safemath{\resB}{\matR_b}	
\safemath{\subD}{\matS} 
\safemath{\subA}{\matS_a} 
\safemath{\subB}{\matS_b} 
\safemath{\dicta}{\matA} 	
\safemath{\dictb}{\matB} 	
\safemath{\hollowS}{H}
\safemath{\hollowA}{H_a}
\safemath{\hollowB}{H_b}
\safemath{\cross}{Z}
\safemath{\coh}{\mu_d}			
\safemath{\coha}{\mu_a}			
\safemath{\cohb}{\mu_b}			
\safemath{\mubs}{\nu}	
\safemath{\cohm}{\mu_m} 
\safemath{\dictset}{\setD}	
\safemath{\dictsetp}{\dictset(\coh,\coha,\cohb)}	
\safemath{\dictsetgen}{\dictset_\text{gen}}
\safemath{\dictsetgenp}{\dictsetgen(\coh)}
\safemath{\dictsetonb}{\dictset_\text{onb}}
\safemath{\dictsetonbp}{\dictsetonb(\coh)}

\safemath{\leftside}{U}
\safemath{\rightsideA}{R_a}
\safemath{\rightsideB}{R_b}

\safemath{\indexS}{\setI_S} 

\safemath{\na}{n_a}			
\safemath{\nb}{n_b}			
\safemath{\coeffa}{p_i}	
\safemath{\coeffb}{q_j}	
\safemath{\seta}{\setP}		
\safemath{\setb}{\setQ}     
\safemath{\setw}{\setW}	
\safemath{\setz}{\setZ}	
\safemath{\cola}{\veca}		
\safemath{\colb}{\vecb}		
\safemath{\cold}{\vecd}		
\safemath{\inputvec}{\vecx} 	
\safemath{\error}{\vece}	
\safemath{\noiseout}{\vecz} 	
\safemath{\inputvecel}{x}
\safemath{\inputveca}{\vecx_a}
\safemath{\inputvecb}{\vecx_b}
\safemath{\outputvec}{\vecy}	
\safemath{\lambdamin}{\lambda_{\mathrm{min}}}

\newcommand{\normone}[1]{\vecnorm{#1}_1}

\safemath{\elltwo}{\ell_2}
\safemath{\ellone}{\ell_1}
\safemath{\ellzero}{\ell_0}
\safemath{\ellinf}{\ell_\infty}
\safemath{\licard}{Z(\coh,\coha,\cohb)}
\safemath{\xsol}{\hat{x}}
\safemath{\xbord}{x_b}		
\safemath{\xstat}{x_s}		
\safemath{\xstatLone}{\tilde{x}_s}
\safemath{\order}{\mathcal{O}} 
\safemath{\scales}{\Theta} 
\safemath{\ones}{\mathbf{1}} 
\safemath{\zeroes}{\mathbf{0}} 
\safemath{\thlone}{\kappa(\coh,\cohb)} 
\safemath{\constoneA}{\delta} 
\safemath{\constoneB}{\epsilon} 
\safemath{\nlarge}{L}				   
\safemath{\sumlarge}{S_\nlarge}
\safemath{\maxlarger}{P_\nlarge}	   
\safemath{\Pzero}{\textrm{P0}}	
\safemath{\Pone}{\textrm{P1}}
\safemath{\vecfir}{\vecw}			 
\safemath{\vecsec}{\vecz}
\safemath{\elvecfir}{w}              
\safemath{\elvecsec}{z}				 
\safemath{\nlargefir}{n}
\safemath{\normout}{\gamma}
\safemath{\auxfun}{h}
\safemath{\supp}{\textrm{supp}}

\safemath{\indexa}{\ell}
\safemath{\indexb}{r}
\safemath{\indexc}{i}
\safemath{\indexd}{j}

\safemath{\project}{P}

\usepackage{url}
\newcommand{\algoname}{SPARFA-Top}

\begin{document}

\title{Joint Topic Modeling and Factor Analysis  of \\[0.1cm] Textual Information and Graded Response Data}
\numberofauthors{1}
\author{
Andrew S. Lan, 
Christoph Studer, Andrew E. Waters,  
Richard G. Baraniuk\\[0.15cm]
       \affaddr{Rice University, TX, USA}\\[0.1cm]
       \email{\{mr.lan,\,studer,\,waters,\,richb\}@sparfa.com}}
\date{today}
\maketitle

\begin{abstract}

\sloppy

Modern machine learning methods are critical to the development of large-scale personalized learning systems that cater directly to the needs of individual learners.
The recently developed SPARse Factor Analysis (SPARFA) framework provides a new statistical model and algorithms for machine learning-based learning analytics, which estimate a learner's knowledge of the latent concepts underlying a domain, and content analytics, which estimate the relationships among a collection of questions and the latent concepts.
SPARFA estimates these quantities given only the binary-valued graded responses to a collection of questions.
In order to better interpret the estimated latent concepts, SPARFA relies on a post-processing step that utilizes user-defined tags (e.g., topics or keywords) available for each question. 
In this paper, we relax the need for user-defined tags by extending SPARFA to jointly process both graded learner responses and the text of each question and its associated answer(s) or other feedback.
Our purely data-driven approach (i) enhances the interpretability of the estimated latent concepts without the need of explicitly generating a set of tags or performing a post-processing step, (ii) improves the prediction performance of SPARFA, and (iii) scales to large test/assessments where human annotation would prove burdensome. 
We demonstrate the efficacy of the proposed approach on two real educational datasets. 

\fussy
\end{abstract}




\section{Introduction}
\label{sec:intro}

Traditional education typically provides a ``one-size-fits-all'' learning experience, regardless of the potentially different backgrounds, abilities, and interests of individual learners. 
Recent advances in machine learning enable the design of computer-based systems that analyze learning data and provide feedback to the individual learner. 
Such an approach has great potential to revolutionize today's education by offering a high-quality, personalized learning experience to learners on a global scale. 

\sloppy

\subsection{Personalized learning systems}

Several efforts have been devoted into building statistical models and algorithms for learner data analysis.
In \cite{ourwork}, we proposed a personalized learning system (PLS) architecture with two main ingredients: (i) \emph{learning analytics} (analyzing learner interaction data with learning materials and questions to provide personalized feedback) and (ii) \emph{content analytics} (analyzing and organizing learning materials including questions and text documents).
We introduced the SPARse Factor Analysis (SPARFA) framework for learning and content analytics, which decomposes assessments into different knowledge components that we call \emph{concepts}. 
SPARFA automatically extracts (i) a question--concept association graph, (ii) learner concept knowledge profiles, and (iii) the intrinsic difficulty of each question, solely from graded binary learner responses to a set of questions; see Fig.~\ref{tbl:a} for an example of a graph extracted by SPARFA.
This framework enables a PLS to provide personalized feedback to learners on their concept knowledge, while also estimating the question--concept relationships that reveal the structure of the underlying knowledge base of a course.

\fussy

The original SPARFA framework~\cite{ourwork} extracts the concept structure of a course from binary-valued question--response data. 
The latent concepts are ``abstract'' in the sense that they are estimated from the data rather than dictated by a subject matter expert.
To make the concepts interpretable by instructors and learners, SPARFA performs an ad-hoc post-processing step to fuse instructor-provided question tags to each estimated concept. 
Requiring domain experts to label the questions with tags is an obvious limitation to the approach, since such tags are often incomplete or inaccurate and thus provide insufficient or unreliable information.

Inspired by the recent success of modern text processing algorithms, such as latent Dirichlet allocation (LDA) \cite{bleilda}, we posit that the text associated with each question can potentially reveal the meaning of the estimated latent concepts without the need of instructor-provided question tags. 
Such an data-driven approach would be advantageous as it would easily scale to domains with thousands of questions.
Furthermore, directly incorporating textual information into the SPARFA statistical model could potentially improve the estimation performance of the approach. 

\sloppy

\subsection{Contributions}

In this paper, we propose \emph{SPARFA-Top}, which extends the SPARFA framework~\cite{ourwork} to jointly analyze both graded learner responses to questions and the text of the question, response, or feedback.
We augment the SPARFA model by statistically modeling the word occurrences associated with the questions as \emph{Poisson} distributed. 
We develop a {computationally efficient} block-coordinate descent algorithm that, given only binary-valued graded response data and associated text, estimates (i) the question--concept associations, (ii) learner concept knowledge profiles, (iii) the intrinsic difficulty of each question, and (iv) a list of most important keywords associated with each estimated concept.

SPARFA-Top is capable of automatically generating a human readable interpretation for each estimated concept in a purely data-driven fashion (i.e., no manual labeling of the questions is required), thus enabling a PLS to automatically recommend remedial or enrichment material to learners that have low/high knowledge level on a given concept. 
Our experiments on real-world educational datasets indicate that SPARFA-Top outperforms SPARFA in terms of both prediction performance and interpretability of the estimated concepts.

\fussy

\subsection{Related work}

\sloppy

The joint analysis of binary-valued data and associated textual information has been studied in the context of congressional voting patterns using Bayesian inference methods \cite{carintop1,carintop}. 
Our proposed approach uses a block-coordinate descent method \cite{wotao} that is computationally more efficient; this aspect is crucial in practice as it enables to provide real-time feedback to each learner. In addition, the particular structure imposed by SPARFA (non-negativity and sparsity) distinguishes our framework from the methods in  \cite{carintop1,carintop}.
Optimization-based topic model methods have been proposed in \cite{ngtop,xing}. None of these methods, however, consider the joint analysis of textual information and other forms of observed data (such as graded learner responses).

\fussy


\section{The SPARFA-Top Model}
\label{sec:model}

\sloppy

We start by summarizing the SPARFA framework~\cite{ourwork}, and then extend it by modeling word counts extracted from textual information available for each question. We then detail the SPARFA-Top algorithm, which jointly analyzes binary-valued graded learner responses to questions as well as question text to generate (i) a question--concept association graph and (ii) keywords for each estimated concept.

\fussy

\sloppy
\subsection{SPARse Factor Analysis (SPARFA)}

 SPARFA  assumes that graded learner response data consist of $N$ learners answering a subset of $Q$ questions that involve $K \ll Q,N$ underlying (latent) concepts.
Let the column vector $\bmc_j \in \mathbb{R}^{K}$, $j \in \{1,\ldots,N\}$ represent the latent \textit{concept knowledge} of the $j^\text{th}$ learner, let $\vecw_i \in \mathbb{R}^{K}$, $i \in \{1,\ldots,Q\}$ represent the \emph{associations} of question $i$ to each concept, and let the scalar $\mu_i \in \mathbb{R}$ represent the \emph{intrinsic difficulty} of question $i$. 
The student--response relationship is modeled as
\begin{align} 
\notag & Z_{i,j} = \vecw_i^T \vecc_j + \mu_i, \quad \forall i,j, \quad \text{and} \\ \quad & Y_{i,j} \sim \textit{Ber}(\Phi(\tau_{i,j}Z_{i,j})),\quad (i,j)\in\Omega_\text{obs},  \label{eq:qa} 
\end{align}
where $Y_{i,j} \in \{0,1\}$ corresponds to the observed binary-valued graded response variable of the $j^\text{th}$ learner to the $i^\text{th}$ question, where~$1$ and $0$ indicate correct and incorrect responses, respectively. 
$\textit{Ber}(z)$ designates a Bernoulli distribution with success probability $z$, and $\Phi(x)= \frac{1}{1+e^{-x}}$ denotes the inverse logit link function, which maps a real value to the success probability $z\in[0,1]$. 
The set $\Omega_\text{obs}$ contains the indices of the observed entries (i.e., the observed data may be incomplete). 

The precision parameter $\tau_{i,j}$ models the \emph{reliability} of the observed binary graded response $Y_{i,j}$. 
Larger values of $\tau_{i,j}$ indicate higher reliability on the observed graded learner responses, while smaller values indicate lower reliability.
For the sake of simplicity of exposition, we will assume \mbox{$\tau_{i,j} = \tau, \, \forall i,j$}, throughout this paper.

To address the fundamental identifiability issue in factor analysis and to account for real-world educational scenarios, \cite{ourwork} imposed specific constraints on the model \fref{eq:qa}. Concretely, every row $\vecw_i$ of the question--concept association matrix $\bW$ is assumed to be \textit{sparse} and \textit{non-negative}. 
The sparsity assumption dictates that one expects each question to be related to only a few concepts, which is typical for most education scenarios. The non-negativity assumption characterizes the fact that knowledge of a particular concept does not hurt one's ability of answering a question correctly.

\sloppy
\subsection{SPARFA-TOP: Joint analysis of learner responses and textual information}

SPARFA~\cite{ourwork} utilizes a post-processing step to link pre-defined tags with the inferred latent concepts.  
We now introduce a novel approach to $\textit{jointly}$ consider graded learner response and associated textual information, in order to directly associate keywords  with the estimated concepts. 

Assume that we observe the word--question occurrence matrix $\bB \in \mathbb{N}^{Q \times V}$, where $V$ corresponds to the size of the vocabulary, i.e., the number of \emph{unique} words that have occurred among the $Q$ questions. Each entry $B_{i,v}$ represents how many times the $v^\text{th}$ word occurs in the associated text of the $i^\text{th}$ question; as is typical in the topic model literature, common stop words (``the'', ``and'', ``in'' etc.) are excluded from the vocabulary.
The word occurrences in $\bB$ are modeled as follows:
\begin{align} 
A_{i,v} = \vecw_i^T \vect_v  \quad \text{and} \quad B_{i,v} \sim \sl{Pois}(A_{i,v}),\,  \quad \forall i,v, \label{eq:qt} 
\end{align}
where $\vect_v \in \mathbb{R}_+^K$ is a non-negative\footnote{Since the Poisson rate $A_{i,v}$ must be strictly positive, we assume that $A_{i,v} \geq \varepsilon$ with $\varepsilon=10^{-6}$ in all experiments.} column vector that characterizes the expression of the $v^\text{th}$ word in every concept.
Inspired by the topic model proposed in \cite{xing}, the entries of the word-occurrence matrix $B_{i,v} $ in \fref{eq:qt} are assumed to be \emph{Poisson} distributed, with rate parameters $A_{i,v}$.

We emphasize that the models \fref{eq:qa} and \fref{eq:qt} share the same question--concept association vector, which implies that the relationships between questions and concepts manifested in the learner responses are assumed to be exactly the same as the question--topic relationships expressed as word co-occurrences. 
Consequently, the question--concept associations generating the associated question text are also sparse and non-negative, coinciding with the standard assumptions made in the topic model literature \cite{bleilda,ftm}.


\section{SPARFA-Top algorithm}
\label{sec:algo}

We now develop the SPARFA-Top algorithm by using block multi-convex optimization, to \emph{jointly} estimate $\bW$, $\bC$, $\boldsymbol{\mu}$, and $\bT=[\bmt_1,\ldots,\bmt_V]$ from the observed student--response matrix~$\bY$ and the word-frequency matrix $\bB$. Specifically, we seek to solve the following optimization problem:
\begin{align} 
\label{eq:prob} 
\notag & \underset{\bW,\bC,\bT \colon  W_{i,k} \geq 0\,\forall i,k, T_{k,v} \geq 0\,\forall k,v} {\text{minimize}} \\ & \textstyle  \underset{(i,j)\in \Omega_\text{obs}}{\sum} \!\!\!\!\!-\log p(Y_{i,j}|\vecw_i^T\vecc_j+\mu_i,\tau)  +  \notag \textstyle  \underset{i,v}{\sum} -\log p(B_{i,v}|\vecw_i^T\vect_v) \\ & \quad\,+ \lambda \textstyle \underset{i}{\sum} \| \vecw_i \|_1 +  \frac{\gamma}{2}  \textstyle \underset{j}{\sum} \| \vecc_j \|_2^2 + \frac{\eta}{2} \textstyle \underset{v}{\sum} \|\vect_v\|_2^2.
\end{align}
Here, the probabilities $p(Y_{i,j}|\vecw_i^T\vecc_j+\mu_i, \tau)$ and $p(B_{i,v} | \vecw_i^T\vect_v)$ follow the statistical models in \fref{eq:qa} and~\fref{eq:qt}, respectively.
The $\ellone$-norm penalty term $\lambda \textstyle \sum_{i} \| \vecw_i \|_1$ induces sparsity on the question--concept matrix~$\bW$. The $\elltwo$-norm penalty terms $\frac{\gamma}{2}  \sum_{j} \| \vecc_j \|_2^2$ and $\frac{\eta}{2} \sum_{v} \|\vect_v\|_2^2$ gauge the norms of the matrices $\bC$ and~$\bT$. 
To simplify the notation, the intrinsic difficulty vector $\boldsymbol{\mu}$ is added as an additional column of $\bW$ and  with~$\bC$ augmented with an additional all-ones row.
The precision parameter $\tau$ serves as a balance between the observed learner responses and question text. For $\tau \rightarrow \infty$, \algoname{} corresponds to SPARFA, while for $\tau \rightarrow 0$, \algoname{} corresponds to topic models.

The optimization problem \fref{eq:prob} is block multi-convex, i.e., the subproblem obtained by holding two of the three factors $\bW$, $\bC$, and $\bT$ fixed and optimizing for the other is convex. This property inspires us to deploy a block coordinate descent approach to compute an approximate to \fref{eq:prob}. The SPARFA-Top algorithm starts by initializing $\bW$, $\bC$, and $\bT$ with random matrices and then optimizes each of these three 
 factors iteratively until convergence. The subproblems of optimizing over $\bW$ and $\bC$ are solved iteratively using  algorithms relying on the FISTA framework (see~\cite{fista} for the details). 

The subproblem of optimizing over $\bC$ with $\bW$ and $\bT$ fixed was detailed in~\cite{ourwork}. The subproblem of optimizing over $\bT$ with $\bW$ and $\bC$ fixed is separable in each column of $\bT$, with the problem for $\vect_v$ being: 
\begin{align} 
\label{eq:solvet} 
 \underset{\vect_v: T_{k,v} \geq 0, \forall k}{\text{minimize}} \quad \textstyle  \sum_i -\log p(B_{i,v}|\vecw_i^T \vect_v) + \frac{\eta}{2} \|\vect_v\|_2^2.
 \end{align}
This subproblem can be efficiently solved using FISTA, where the gradient of the objective function with respect to $\vect_v$ being:
\begin{align}
\nabla_{\vect_v} \textstyle  \sum_i -\log p(B_{i,v}|\vecw_i^T\vect_v) + \frac{\eta}{2} \|\vect_v\|_2^2 = \bW^T \vecr + \eta\,\vect_v,
\end{align}
where $\vecr$ is a $Q \times 1$ vector with its $i^\text{th}$ element being $r_i = 1 - \frac{B_{i,v}}{\vecw_i^T \vect_v}$. 
The projection step corresponds to the simple operation of setting negative entries in $\vect_v$ to zero.

The subproblem of optimizing over $\bW$ with $\bC$ and $\bT$ fixed is also separable in each row of $\bW$. The problem for each $\vecw_i$ is:
\begin{align}
\notag \underset{\vecw_i\colon W_{i,j}\geq0 \forall j}{\text{minimize}} \quad & \textstyle  \sum_{j: (i,j)\in \Omega_\text{obs}} -\log p(Y_{i,j}|\vecw_i^T \vecc_j+\mu_i,\tau) \\ & +  \textstyle  \sum_v -\log p(B_{i,v}|\vecw_i^T \vect_v) + \lambda \normone{\vecw_i},
\end{align}
which can be efficiently solved using FISTA. Specifically, analogous to \cite[Eq.~5]{ourwork}, the gradient of the smooth part of the objective function with respect to $\vecw_i$ corresponds to:
\begin{align}
\notag \nabla_{\vecw_i} & \big( \textstyle \sum_{j: (i,j)\in \Omega_\text{obs}} -\log p(Y_{i,j}|\vecw_i^T \vecc_j+\mu_i,\tau) \\ &+ \textstyle \! \sum_v \! -\log p(B_{i,v}|\vecw_i^T \vect_v) \big) \! = \! - \bC^T (\vecy_i - \vecp) + \bT^T \vecs,
\end{align}
where $\vecy_i$ represents the transpose of the $i^\text{th}$ row of $\bY$, $\vecp$ represents a $N \times 1$ vector with $p_j =1/(1+e^{-\vecw_i^T \vecc_j})$ as its $j^\text{th}$ element, and $\vecs$ is a $V \times 1$ vector with $s_v = 1 - \frac{B_{i,v}}{\vecw_i^T \vect_v}$ as its $v^\text{th}$ element. The projection step is a soft-thresholding operation, as detailed in \cite[Eq.~7]{ourwork}. The step-sizes are chosen via back-tracking line search as described in \cite{boydbook}.

Note that we treat $\tau$ as a fixed parameter. Alternatively, one could estimate this parameter \emph{within} the algorithm by introducing an additional step that optimizes over $\tau$. 
A throughout analysis of this approach is left for future work.

\sloppy
\section{Experiments}
\label{sec:real}

\begin{figure}
\vspace{-0.5cm}
\includegraphics[width=.84\columnwidth]{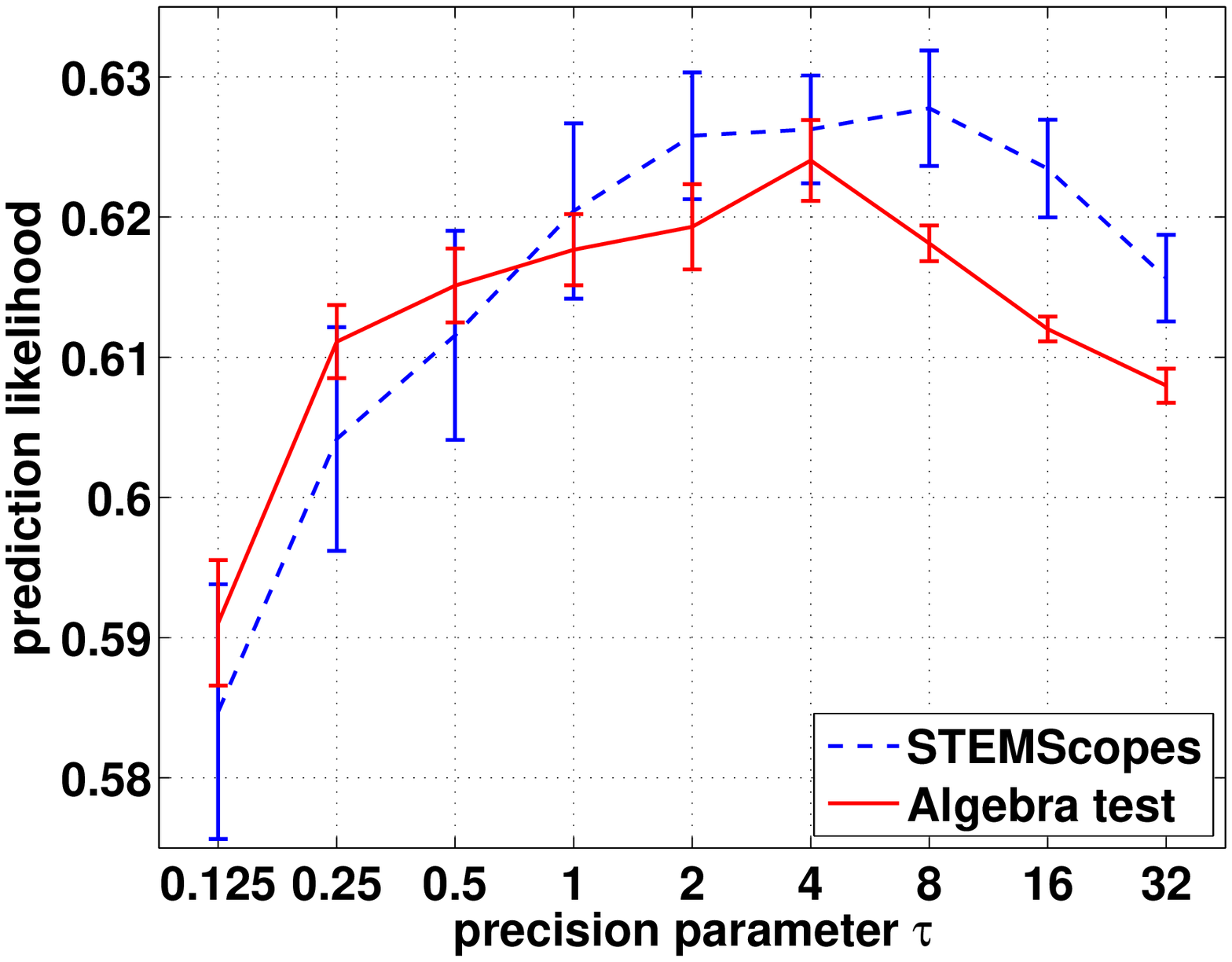}
\caption{Average predicted likelihood on 20\% hold-out data in $\bY$ using \algoname{} with different precision parameters $\tau$. For $\tau\to\infty$ SPARFA-Top corresponds to SPARFA as proposed in \cite{ourwork}.\label{fig:curve}}
\vspace{-0.5cm}
\end{figure}

\begin{figure*}[t]
\vspace{-0.3cm}
\begin{floatrow}
\ffigbox{%
\includegraphics[width=.82\columnwidth]{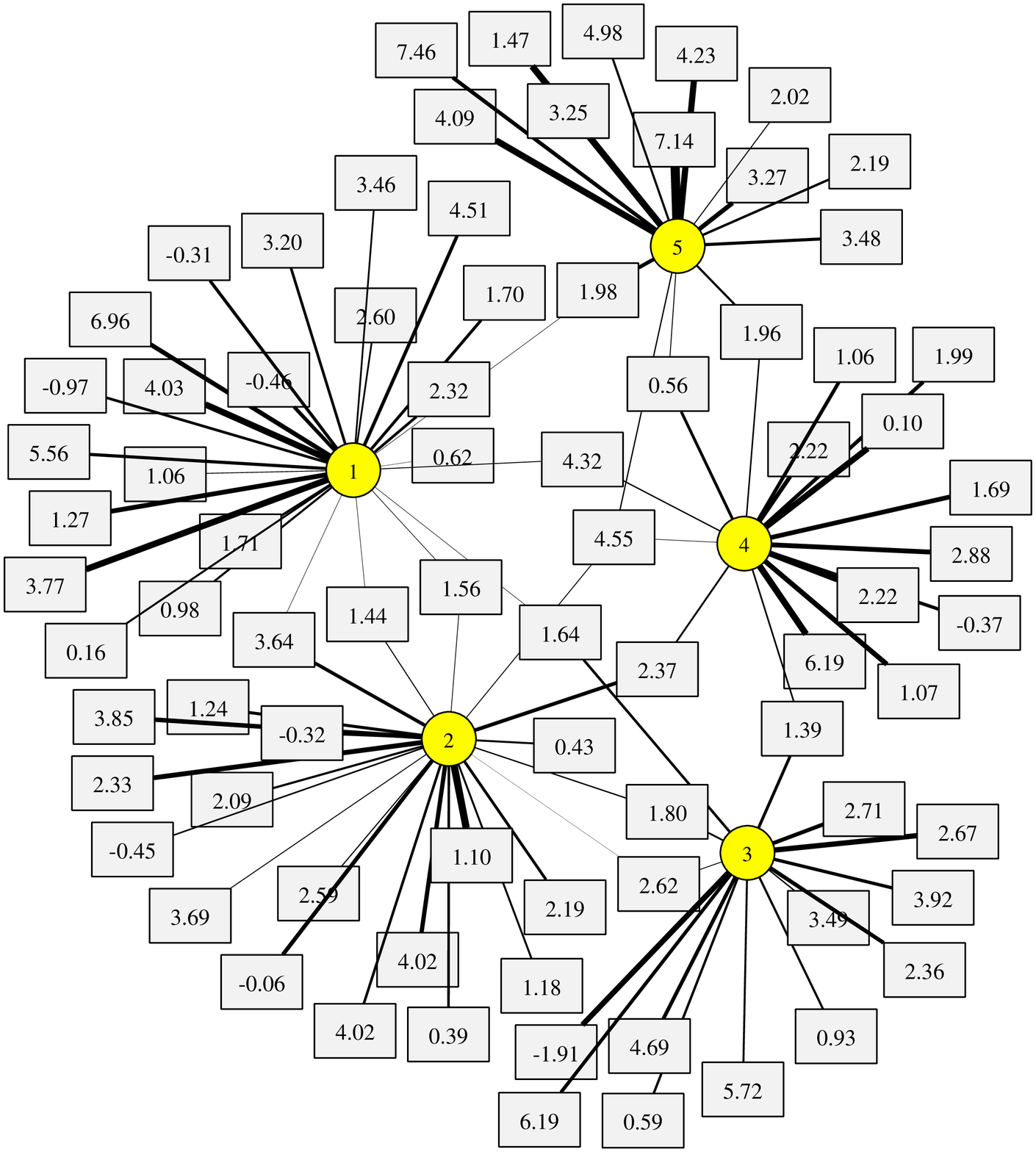} \\[0.2cm]
\scalebox{.78}{%
\begin{tabular}{lll}
\toprule[0.2em]
Concept 1 & Concept 2 & Concept 3\\
\midrule[0.1em]
Energy & Water & Plants \\
Water & Percentage & Buffalo \\ 
Earth & Sand & Eat \\
\midrule[0.2em]
 Concept 4 & Concept 5\\ 
\midrule[0.1em]
Water & Water \\
Soil & Heat \\ 
Sample & Objects \\
\bottomrule[0.2em]\\
\end{tabular}}
\vspace{-0.8cm}
}{\caption{Question--concept association graph and most important keywords recovered by \algoname{} for the STEMscopes dataset; boxes represent questions, circles represent concepts, and thick lines represent strong question--concept associations.} \label{tbl:a}}
\ffigbox{%
\vspace{0.2cm}
\includegraphics[width=.82\columnwidth]{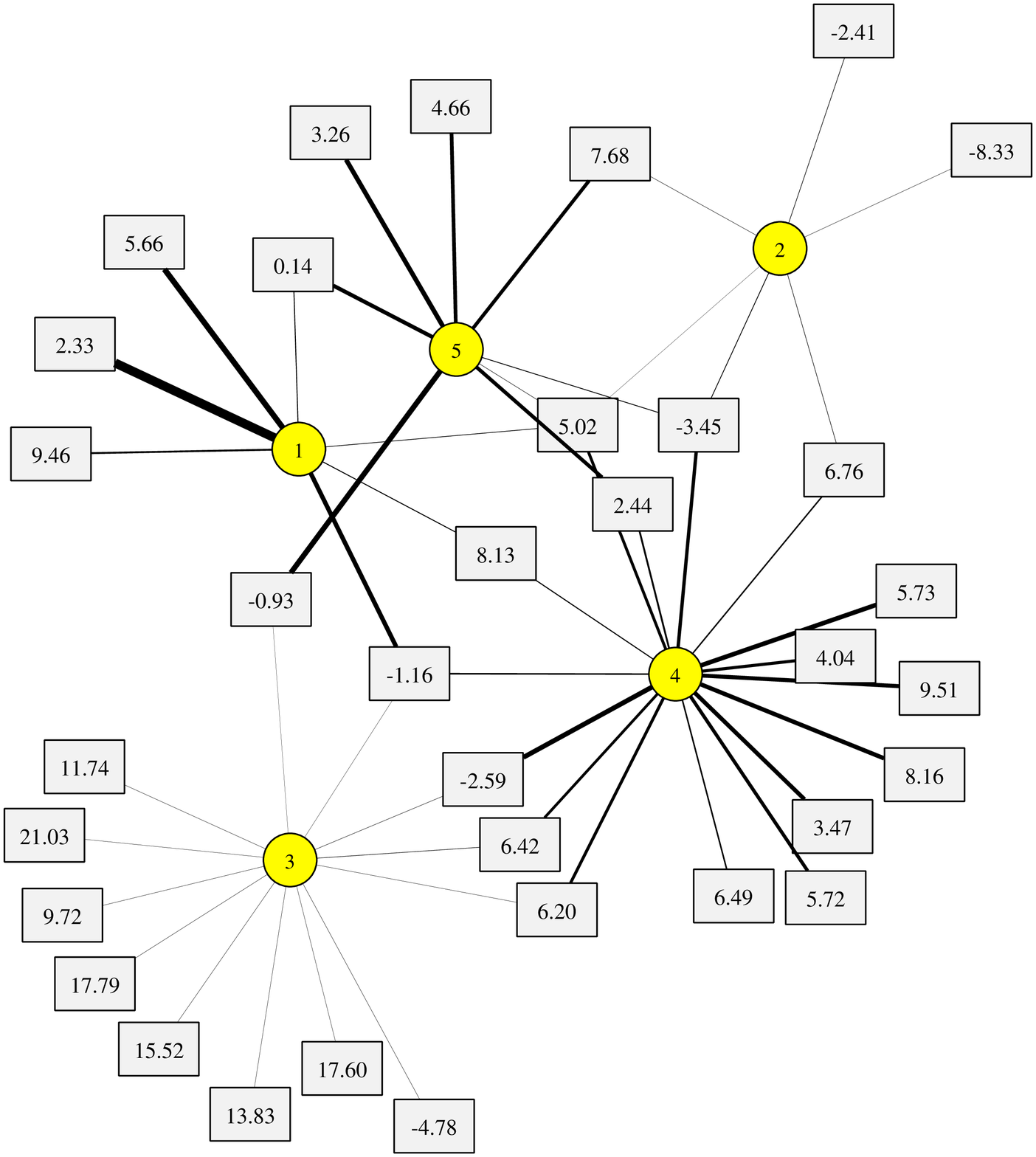} \\[0.0cm]
\scalebox{.8}{%
\begin{tabular}{lll}
\toprule[0.2em]
Concept 1 & Concept 2 & Concept 3\\
\midrule[0.1em]
Solving equations & Slope & Arithmetic \\
Quadratic function & Fractions & Trigonometry \\ 
Fractions & Simplifying expressions & System of equations \\
\midrule[0.2em]
 Concept 4 & Concept 5\\ 
\midrule[0.1em]
Simplifying expressions & Inequality \\
Factoring polynomials & Plotting functions \\ 
Fractions & Geometry \\
\bottomrule[0.2em]\\
\end{tabular}}
\vspace{-0.7cm}
}{\caption{Question--concept association graph and~3 most important keyword recovered by \algoname{} for the algebra test dataset; boxes represent questions, circles represent concepts, and thick lines represent strong question--concept associations.} \label{tbl:b}}
\end{floatrow}
\end{figure*}

We now demonstrate the efficacy of \algoname{} on two real-world educational datasets: an $8^{\text{th}}$ grade Earth science course dataset provided by STEMscopes \cite{stemwebsite} and a high-school algebra test dataset administered on Amazon's Mechanical Turk \cite{mechturkwebsite}, a crowdsourcing marketplace.
The STEMscopes dataset consists of 145 learners answering 80 questions, with only 13.5$\%$ of the total question/answer pairs being observed. The question--associated text vocabulary consists of 326 words, excluding common stop-words. 
The algebra test dataset consists of 99 users answering 34 questions, with the question--answer pairs fully observed. As no informative question text is available, we use the tags on each question to from a vocabulary of 13 words.

The regularization parameters $\lambda$, $\gamma$ and $\eta$, together with the precision parameter $\tau$ of SPARFA-Top, are selected via cross-validation.
In \fref{fig:curve}, we show the prediction likelihood defined by $p(Y_{i,j} | \vecw_i^T\vecc_j+\mu_i,\tau), (i,j) \in \bar{\Omega}_\text{obs}$ for \algoname{} on 20\% holdout entries in $\bY$ and for varying precision values~$\tau$. We see that textual information can slightly improve the prediction performance of SPARFA-Top over SPARFA (which corresponds to $\tau \to \infty$), for both the STEMscopes dataset and the algebra test dataset.
The reason for (albeit slight) improvement in prediction performance is the fact that textual information reveals additional structure underlying a given test/assessment.

Figures \ref{tbl:a} and \ref{tbl:b} show the question--concept association graphs along with the recovered intrinsic difficulties, as well as the top three words characterizing each concept, for both datasets. 
Compared to SPARFA (see~\cite{ourwork}), we observe that \algoname{} is able to relate all questions to concepts, including those questions that were found  in~\cite[Figs.~2 and~9]{ourwork} to be unrelated to any concept.
Furthermore, Figures~\ref{tbl:a} and~\ref{tbl:b} demonstrate that \algoname{} is capable of automatically generating an interpretable summary of the meaning of each concept.
%


\section{Conclusions}
\label{sec:conclusions}

\sloppy

We have introduced the \algoname{} framework, which extends SPARFA~\cite{ourwork} by jointly analyzing both the binary-valued graded learner responses to a set of questions and the text associated with each question via a topic model. 
As our experiments have shown, our purely data-driven approach avoids the manual assignment of tags to each question and significantly improves the interpretability of the estimated concepts by automatically associating keywords extracted from question text to each estimated concept. 
\fussy


{ \small
\bibliographystyle{abbrv}
\bibliography{sparfaclustbib}
}

\end{document}